%% file: conference.tex
\documentclass{article} 
\usepackage{conference,times}

\input{math_commands.tex}

\usepackage{booktabs} 
\usepackage{hyperref}
\usepackage{url}
\usepackage{algorithm}
\usepackage{algpseudocode}
\usepackage{algpseudocode}
\usepackage{amsmath}
\usepackage{graphicx}

\finalcopy
\title{Adaptive Test-Time Reasoning via Reward-Guided Dual-Phase Search}

\author{\\
\textbf{~~ Yingqian Cui$^{1,2}$\thanks{{Work done during her internship at Amazon.}}   ~~ Zhenwei Dai$^{2}$ ~~ Pengfei He$^{1}$ ~~ Bing He$^{2}$ ~~ Hui Liu$^{2}$ ~~ Xianfeng Tang$^{2}$  } \\ 
\textbf{~~~~ ~~Jingying Zeng$^{2}$ ~~ Suhang Wang$^{3}$~~ Yue Xing$^{1}$ ~~ Jiliang Tang$^{1}$ ~~ Benoit Dumoulin $^{2}$ }\\
 ~~ ~~ ~~~ ~~~ ~~ $^{1}$Michigan State University
 ~~$^{2}$ Amazon ~~ $^{3}$ Pennsylvania State University \\}

\begin{document}

\maketitle

\begin{abstract}
Large Language Models (LLMs) have achieved significant advances in reasoning tasks. A key approach is tree-based search with verifiers, which expand candidate reasoning paths and use reward models to guide pruning and selection. 
Although effective in improving accuracy, these methods are not optimal in terms of efficiency: they perform simple decomposition on the reasoning process, but ignore the planning-execution nature of tasks such as math reasoning or code generation. This results in inefficient exploration of reasoning process.
To address this, we propose a dual-phase test-time scaling framework that explicitly separates reasoning into planning and execution, and performs search over the two phases individually.
Specifically, we decompose reasoning trajectories and develop reward models for each phase, enabling the search to explore and prune plans and executions separately. 
We further introduce a dynamic budget allocation mechanism that adaptively redistributes sampling effort based on reward feedback, allowing early stopping on confident steps and reallocation of computation to more challenging parts of the reasoning process. Experiments on both mathematical reasoning and code generation benchmarks demonstrate that our approach consistently improves accuracy while reducing redundant computation.

\end{abstract}

\vspace{-0.05in}
\section{Introduction}\label{sec:intro}
\vspace{-0.06in}
In recent years, Large Language Models (LLMs) have achieved remarkable success in complex reasoning tasks such as mathematical problem solving, code generation, and decision making~\citep{chen2021evaluating,yao2023react}. A common approach to improve LLM reasoning is Chain-of-Thought (CoT) prompting~\citep{wei2022chain}, which guides the model to generate intermediate reasoning steps in a stepwise manner, often improving performance on arithmetic and symbolic tasks. 


To further enhance the quality and accuracy of multi-step reasoning, recent works have explored test-time scaling methods, which perform structured search or sampling over multiple reasoning paths. 
According to~\cite{snell2024scaling}, these methods consider two main directions: (1) \textbf{Distribution-Based Sampling}, by refining how candidates are generated, e.g., through iterative revision~\citep{muennighoff2025s1,shinn2023reflexion} or parallel sampling with selection~\citep{snell2024scaling, diao2023active}; and (2) \textbf{Reward-Based Searching}, by using verifiers or reward models to select or guide promising reasoning paths~\citep{snell2024scaling, wu2024inference}. Among the latter, Process Reward Modeling (PRM)~\citep{wu2024inference} has shown strong performance by evaluating partial reasoning steps and guiding the search process accordingly. {Instead of judging only final outcomes, PRM provides fine-grained supervision at the process level, assigning rewards to intermediate steps and enabling the model to distinguish between useful and unproductive reasoning trajectories when generating the intermediate steps. This step-level feedback allows search algorithms to prune low-quality candidates earlier and concentrate computation on promising directions, leading to more efficient and accurate reasoning.}
%
While existing PRM-based test-time scaling methods have significantly improved the reasoning performance of LLMs, several key limitations remain. 

First, {existing literature on test-time scaling methods scales up the computation based on simple decomposition of the whole reasoning process, and there is limited understanding on how test-time scaling behaves beyond the simple decomposition. In particular, }complex tasks such as mathematical problem solving and code generation naturally involve two distinct cognitive phases: planning, which entails high-level strategic formulation (e.g., ``define variables and set up an equation''), and execution, which involves carrying out precise computations or implementations (e.g., arithmetic calculations or code writing)~\citep{zhou2022least, wang2024planning, hao2023reasoning, wang2023plan}.
Although the benefits of explicitly writing out plans during reasoning have been discussed~\citep{zhou2022least, wang2024planning, hao2023reasoning,wang2023plan}, most literature in test-time scaling treats planning and execution as a unified pipeline: the model {generates a plan immediately followed by its execution}, and both are evaluated together.
\begin{figure}
    \centering
    \includegraphics[width=0.85\linewidth]{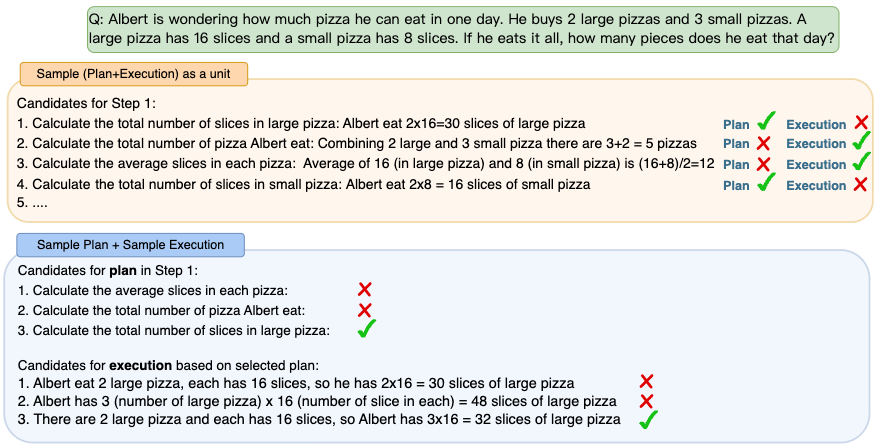}
    \vspace{-0.1in}
    \caption{\small{An example of reasoning with plan and execution as a single unit versus searched separately. }}
    \label{fig:whysplit}
    \vspace{-0.2in}
\end{figure}
The consequence is that, if a step has a correct plan but an incorrect execution, the entire candidate is discarded, wasting the useful partial structure. Conversely, if a step is already flawed at the planning stage, the search still wastes budget generating its executions. For example, consider the problem in Figure~\ref{fig:whysplit}. Since both the plan and the execution may contain errors, sampling them as a unit often requires many trials before obtaining a step in which both are correct simultaneously.




The second limitation of existing tree-based test-time scaling methods is that they mainly adopt a fixed sampling budget per step (e.g., sampling $k$ candidates at every reasoning step), ignoring the varying difficulty across different steps within the same question. This rigid allocation can lead to inefficient computation, especially when simple steps receive excessive attention while more challenging parts remain underexplored. {While there are some studies that explore sample-wise budget allocation, i.e., dynamically distributing the overall budget across different questions or different candidate trajectories~\citep{zuo2025strategic, lin2025plan}, these approaches do not address the problem of step-wise allocation within a single reasoning trajectory.}


To address these limitations, we propose {\emph{DREAM, a \underline{D}ual-phase \underline{RE}ward-guided \underline{A}daptive reasoning framework at test ti\underline{M}e}}.
Unlike prior methods that treat each plan–execution pair as a single unit, DREAM explicitly conducts search in two stages: it first searches over multiple planning candidates and uses a reward model to select promising subgoals, and then, conditioned on the selected plans, it searches over execution candidates and applies a second reward evaluation to retain the most reliable solutions. 
For example, for the question in Figure~\ref{fig:whysplit}, we first sample candidate plans and select the promising ones. Then, conditioned on these plans, we generate multiple execution candidates. This two-stage procedure ensures that poor plans are eliminated early, while promising plans can be paired with different execution attempts until the correct result is found, which ensure that computation is allocated more efficiently across the two phases.
In addition to the above, we further incorporates DREAM with a \emph{dynamic budget allocation} mechanism that adaptively adjusts the number of samples at both phases based on real-time reward feedback, enabling early stopping on easy steps and reallocating resources to harder ones.

To verify the effectiveness of the proposed algorithm, we conduct comprehensive evaluations across two domains: math reasoning and code generation. Experimental results show that our approach not only improves answer accuracy but also enhances test-time efficiency. 
\vspace{-0.1in}
\section{Related works}
\vspace{-0.06in}
\textbf{Test-time Scaling.}
Test-time scaling methods improve reasoning quality without parameter updates by expending more computation at inference.
According to~\cite{snell2024scaling}, two primary mechanisms for scaling test-time include (1) {Distribution-Based Sampling} and (2) {Reward-Based Searching}.
Methods of (1) include {s1}~\citep{muennighoff2025s1} and Reflexion~\citep{shinn2023reflexion}, which introduces sequential self-revision to iteratively refine candidate solutions, and {Best-of-N}~\citep{wang2022self}, which samples multiple candidate reasoning chains in parallel and aggregates via majority voting (self-consistency). 

Methods of (2) work by treating intermediate reasoning states as nodes in a search tree and expand continuations via the base LLM. They include MCTS-based methods, such as RAP~\citep{hao2023reasoning}, LiteSearch~\citep{wang2024litesearch}, rStar~\citep{qi2024mutual} and rStar-Math~\citep{guan2025rstar}, which apply Monte Carlo Tree Search to explore reasoning paths, and verifier-based methods, which rely on outcome-level judges~\citep{cobbe2021training, snell2024scaling} or process-supervised reward
models (PRMs)~\citep{lightman2023let,wu2024inference,hooper2025ets} to score and prune candidates. 
Moreover, \citet{setlur2025scaling} indicate that verifier-based methods combined with with search-based strategy are provably better than verifier-free approaches. While effective, most current approaches (even those that adopt a plan-execution format) still treat planning and execution as a single unified process, without performing separate search or adaptive budget allocation across the two phases. {Moreover, we would like to highlight that, although a variety of test-time methods have been proposed, in our experiments we mainly consider reward-model-based methods as baselines to ensure a fair comparison, as reward models provide additional information beyond the base LLM.}

\textbf{Code Generation with LLMs.}
Recent work has explored diverse strategies (including test-time scaling) to enhance LLMs for code generation. For example, $\text{S}^*$\citep{li2025s} employs parallel sampling with sequential scaling and adaptive input synthesis to improve code generation. \cite{yu2025z1} introduce Z1, which trains the LLM on both short and long reasoning trajectories and leverages a shifted thinking window to enable the model to adaptively control the length of its ‘thinking’ process according to problem complexity. In addition, PlanSearch~\citep{wang2024planning} boosts code generation by exploring diverse natural-language plans before translating them into code.
In addition, tree-structured or agent-based searching frameworks like CodeTree~\citep{li2024codetree}, Tree-of-Code~\citep{ni2024tree} and Funcoder~\cite{chen2024divide} design stepwise generation or refinement algorithms for code generation, where candidate programs are expanded or revised through structured search, demonstrating the benefits of structured exploration. 

Although many of the above code generation methods also introduce a planning stage before the generation of code, they typically focus only on improving or selecting a good plan to guide execution rather than performing separate search processes for planning and execution. In contrast, our work performs dual-phase scaling and selection, and assigns budget across phases, which has the potential of having a more effective use of computation.
\vspace{-0.1in}
\section{Proposed Method}
\vspace{-0.06in}
In this section, we present the general idea and design details of our proposed method. 
In Section~\ref{mth:idea}, we introduce the dual-phase search over planning and execution, 
which serves as the foundation of our framework. {We then utilize math reasoning task and code generation task as examples to implement the main idea of the dual-phase search, and introduce the main algorithms for both tasks.}
In Section~\ref{mth:reward}, we describe how to develop the reward models, 
including both the construction of training data and the design of the reward function.

\begin{figure}
    \centering
    \vspace{-0.25in}
    \includegraphics[width=0.95\linewidth]{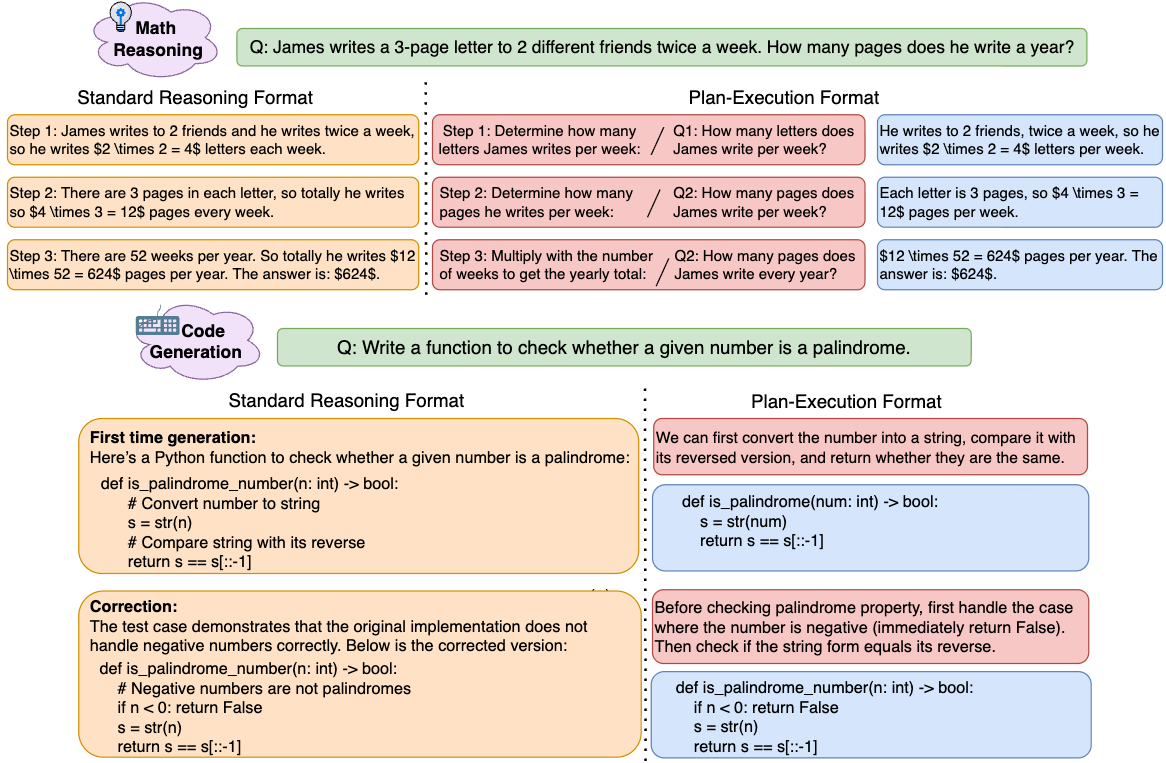}
     \vspace{-0.06in}
    \caption{\small{Plan–execution Format for Reasoning in Math and Code Tasks.} }
    \label{fig:format}
\end{figure}

\vspace{-0.06in}
\subsection{Dual-phase search over planning and execution}\label{mth:idea}

\vspace{-0.06in}
{To introduce the algorithm, we first define the plan-execution reasoning format. Since the detailed implementation of test-time scaling varies among different type of tasks, following other literature in test-time scaling, e.g., \citep{hao2023reasoning, li2024codetree}, we use two representative tasks, math reasoning and code generation, to present our algorithm. }



\textbf{Plan-Execution Reasoning Formats.}
As shown in Figure~\ref{fig:format}, compared with the standard reasoning format where each step directly produces a continuation, in  \emph{plan–execution format}, each step is decomposed into (i) a \emph{plan}, which formulates a sub-question or a subgoal, and (ii) an \emph{execution}, which directly answers the sub-question or completes the subgoal through concrete calculations or derivations. {When the plan is expressed as a sub-question and the execution as its answer, it is known as the \emph{least-to-most} prompting paradigm~\citep{zhou2022least}.} To guide models to follow this format, we provide few-shot examples in the prompts, allowing LLMs to generate solutions in the plan–execution format.





\begin{figure}
\vspace{-0.06in}
    \centering
    \includegraphics[width=0.96\linewidth]{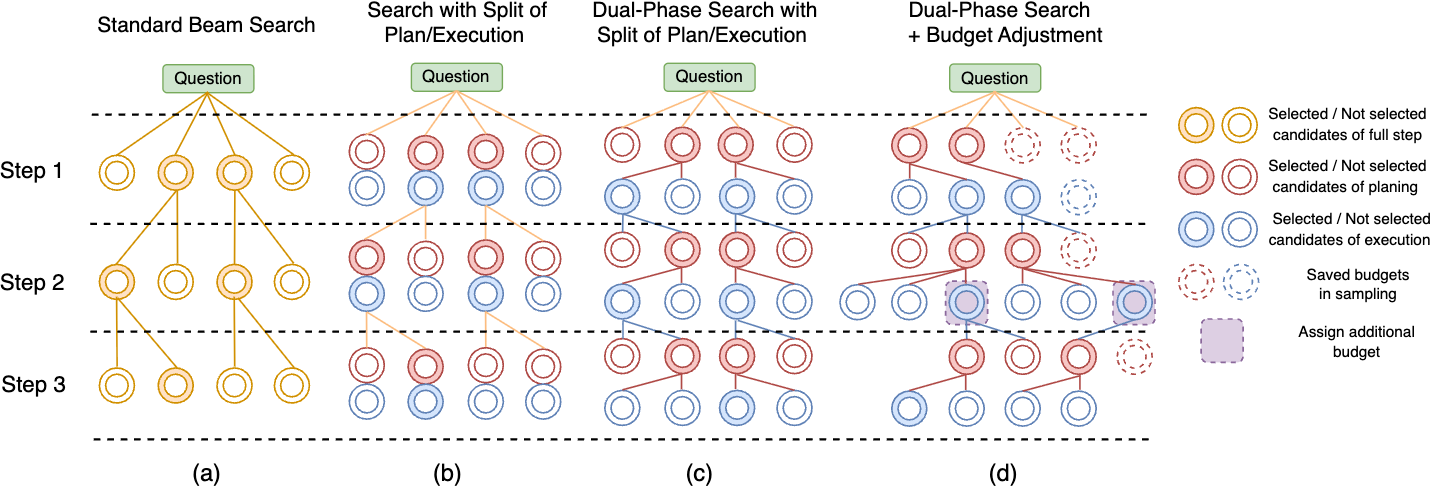}
    \vspace{-0.15in}
    \caption{\small{Workflow of Standard Beam Search and Dual-Phase Search (with budget allocation). }}
    \vspace{-0.2in}
    \label{fig:workflow}
\end{figure}

\textbf{Dual-phase Search for Math Reasoning.}
{Our dual-phase search builds on the standard beam search framework, whose workflow is shown in Figure~\ref{fig:workflow} (a). In standard beam search, we utilize the standard reasoning format, where planning and execution are not explicitly expressed, and use the reasoning model to sample a fixed number of candidates. Reward models are applied to score all candidates, and the top-ranked ones are retained for expansion in the next step. Motivated by prior work highlighting the benefits of explicitly separating planning and execution \citep{zhou2022least, wang2024planning,wang2023plan}, we further extend standard beam search into a variant that outputs plan–execution pairs in a single step, as illustrated in Figure~\ref{fig:workflow}(b). }

{However, as mentioned in Section~\ref{sec:intro}, the design of treating planning and execution as a single unit is inefficient, since it fails to allocate computation appropriately across the two phases. To overcome this drawback, our dual-phase search (Figure~\ref{fig:workflow}(c))}
 explicitly separates each step into a planning phase (red nodes) and an execution phase (blue nodes). In the planning phase, $N_1$ candidate subgoals are sampled and scored by a planning reward model ($\text{PRM}_{\text{plan}}$), where the top $n_1$ candidates are selected. In the execution phase, $N_2$ continuations are generated conditional on the chosen plans and scored by an execution reward model ($\text{PRM}_{\text{exec}}$). Then the top $n_2$ candidates will be selected for the expansion of next step. This separation ensures that the weaker plans can be prune early while promising plans can be given multiple execution attempts, which reduces the risk of discarding good strategies due to execution errors.

In addition, we further extend our method to a \emph{budget-adjusted dual-phase search} that incorporates adaptive allocation. We design this mechanism based on the observation that reasoning difficulty exhibits substantial variance: not only across different problems within a dataset, but also across different steps within the same problem (see examples in Appendix~\ref{appd:examp}). As a result, allocating the same sampling budget to every step of every example is inefficient: easy steps waste resources, while difficult steps remain underexplored.
To address this limitation, we introduce an adaptive allocation strategy that stops early when confident candidates are already found and reallocates additional computation to more challenging steps, improving the overall accuracy–efficiency trade-off.
The specific workflow is shown in Figure~\ref{fig:workflow} (d) and the detailed algorithm can be found in Algorithm~\ref{alg} in Appendix~\ref{appd:alg}. At each step, sampling in both the planning and execution phases follows a two-threshold rule. Specifically, as candidates are sampled and scored, if at least $n_1$ (for planning) or $n_2$ (for execution) candidates {exceed a specific threshold $\tau_{p1}$ (for planning) or $\tau_{e1}$ (for execution), sampling is terminated early without consuming the full budget.}
Conversely, if after exhausting the full budget there is no candidate whose reward value higher than a lower threshold ($\tau_{p2}$/$\tau_{e2}$),\footnote{{In practice, for simplicity, we set $\tau_{p1}$=$\tau_{e1}$ and $\tau_{p2}$=$\tau_{e2}$}} we allow at most an additional $m_1$ (planning) or $m_2$ (execution) samples to be generated in order to search harder steps more thoroughly. This mechanism prevents wasted computation on easy steps with confident high-reward candidates, while allocating extra exploration to uncertain or challenging steps. By combining dual-phase scoring with this adaptive budget policy, our method aligns computation with step-level difficulty and improves efficiency over standard beam search (as will be demonstrated in the experiments in Section~\ref{sec:main}). {Furthermore, as will be discussed in Appendix~\ref{appd:synergy}, the combination of dual-phase search and dynamic budget allocation has a synergy effect, as the two components mutually reinforce each other by reducing wasted computation and reallocating resources to harder steps.}



\textbf{Dual-phase Search for Code Generation.} For code generation tasks, we 
follow the framework of CodeTree~\citep{li2024codetree} {and extend it with our} dual-phase search. {We adopt this framework because its tree-based structure provides a natural backbone for implementing our dual-phase design.} The key difference from math reasoning is that, in code generation task, a visible test set is available for debugging. Instead of treating each step as a partial components of the solution, in CodeTree, each step produces a complete program, and subsequent steps perform iterative debugging based on execution results from failed test cases of earlier solutions.

In the original CodeTree algorithm, both the initial generation and subsequent debugging involve sampling multiple planning candidates per step, which are then attempted sequentially. Whether to expand the current step or backtrack to alternative candidates depends on whether the current node’s reward exceeds that of the previous one. The node's reward value is computed by combining two factors: (i) the percentage of passed test cases, and (ii) a score given by an LLM-based critic agent.

We modify the above framework in several ways to implement the dual-phase search. First, for each step, we apply a dedicated reward model to the $N_1$ sampled planning candidates, rank them, and prioritize higher-scoring plans for execution. Second, in the execution phase, we scale generation by producing $N_2$ candidate solutions conditioned on the chosen plan, and select the one with the highest reward. Third, as evidenced by experimental results {shown in Appendix~\ref{appd:critic}}, the critic agent provided only marginal benefit. As a result, we remove this component and rely solely on the percentage of passed test cases as the execution reward. Finally, we incorporate a budget-adjustment criterion similar to that in mathematical reasoning: if the reward of a generated candidate exceeds a threshold, we stop further sampling and save the unused budget; if the rewards of all sampled candidates fall below another threshold, additional budget is allocated to generate more candidates. Through these modifications, we extend CodeTree into a dual-phase search framework that conducts separate searches for planning and execution, integrates reward methods for each phase, and allocates computation adaptively based on step-level difficulty.


{\textbf{Remark.} We note that the general idea of dual-phase search is not limited to math or code reasoning, but can generalize to a wide range of reasoning tasks, as long as the solutions can be expressed in a plan–execution format and the reasoning process can be organized within a tree-based framework.}

\vspace{-0.06in}
\subsection{Construction of the Reward Models}
\label{mth:reward}

\vspace{-0.06in}
\textbf{Training data.} 
To develop the reward models, we follow the general idea of ~\cite{wang2023math} to construct datasets that evaluates the quality of both planning and execution at each intermediate reasoning step. We begin by generating complete multi-step reasoning trajectories. For each question in the training set, we sample multiple trajectories at a higher decoding temperature to ensure diversity. Each trajectory is expressed as a sequence of step-wise plan–execution pairs that progressively lead toward the final solution.

To annotate the steps, we adopt a rollout-based labeling strategy. 
{For each intermediate plan or execution, we generate five independent continuations beginning from that step using the same LLM that produced the trajectory.}
If at least one of these rollouts leads to a correct final answer, the current step is labeled as positive (“+”); otherwise negative (“–”). This approach assesses the utility of a plan/exeuction based on its downstream impact on solving the problem.

 \textbf{Reward function:} 
 We implement the reward model by fine-tuning an instruction-tuned LLM. The input to the reward model, denoted as $x$, consists of the original question, all preceding reasoning steps (including both plans and executions), and the current plan or execution to be evaluated. In terms of the output, instead of adding a separate classification head, we follow~\cite{dong2024rlhf} to reformulate the prediction as a next-token prediction task: the final position of the input sequence is reserved for a binary label, and the model is trained to output either {``+"} or {``–"} at that position. 
 The reward function is then given as:
 \vspace{-0.1in}
\[
\text{Reward}(x) = \text{softmax}(\ell(x))_{+} = \frac{\exp\left(\ell_{+}(x)\right)}{\exp\left(\ell_{+}(x)\right) + \exp\left(\ell_{-}(x)\right)} 
\] 
where
\( \ell_{+}(x) \)/\( \ell_{-}(x) \) {are the logits output by the model when predicting the special tokens {``+"}/{``–"}.} 




\vspace{-0.06in}
\section{Experiments}
\vspace{-0.06in}
\subsection{Setup}\label{sec:setup}
\vspace{-0.06in}
We evaluate our method in both math-reasoning and code generation tasks. In this subsection, we introduce the experiment setups for both tasks.

\textbf{Maths Reasoning.}
We evaluate our method on two widely used math reasoning benchmarks: GSM8K~\citep{cobbe2021training} and MATH~\citep{hendrycks2021measuring}. For GSM8K, the full training set of approximately 7.5k problems is used to construct training data for the reward model, while evaluation is performed on the 1.3k test set. For MATH, we leverage the 12.5k training problems to build reward-model training data and conduct evalulation on the standard MATH500 benchmark, a representative subset of 500 problems from the MATH test set.
We generate large-scale synthetic trajectories to build up the training dataset: about 400k samples for GSM8K and 400k samples for MATH. The training trajectories and the rollout process for label assignment are produced using LLaMA-3-8B-Instruct (for GSM8K) and Qwen-2.5-3B-Instruct (for MATH).
{We then combine data from both datasets to train the reward models, fine-tuning Qwen-2.5-32B-Instruct~\citep{qwen2.5}. The resulting reward model is applied in experiments on both benchmarks.}

We compare our method with three baselines: majority vote, standard beam search, and REBASE~\citep{wu2024inference}, a tree-based search method that does not implement dual-phase search. To ensure a fair comparison, we format all reasoning in the plan–execution style and use the same reward model (trained with the rollout-based annotation strategy) across all methods which involve using reward models.
For evaluation, we use three different LLMs on each benchmark: Qwen-2.5-MATH-1.5B-Instruct~\citep{yang2024qwen2}, DeepSeekMath-7B-Instruct~\cite{shao2024deepseekmath}, and LLaMA-3-8B-Instruct for GSM8K / LLaMA-3.1-8B-Instruct~\citep{grattafiori2024llama3} for MATH. {We use different versions of LLaMA for the GSM8K and MATH experiments to ensure that the reasoning models have moderate ability relative to the difficulty of each dataset. This choice allows us to better demonstrate the effectiveness of test-time scaling, since improvements are more evident when the base model is neither too strong nor too weak.}


\textbf{Code Generation Reasoning.} 
For code generation, we conduct experiments on HumanEval~\citep{chen2021evaluating} and MBPP~\citep{austin2021program}, including their extended versions (HumanEval+ and MBPP+)~\citep{evalplus}, which include more challenging test cases. The reward model training data is drawn from the MBPP training set (approximately 600 examples), augmented with around 3,000 examples from CodeAlpaca~\citep{codealpaca}. The generation of training trajectories and the labeling process are carried out using a group of Qwen2.5-Coder models ranging in size from 1.5B to 32B. {We use models with different capacities to produce a wider range of trajectories (both correct and incorrect), which improves the diversity of supervision for training the reward model.}
The final reward model is obtained by fine-tuning Qwen-2.5-Coder-7B-Instruct~\citep{hui2024qwen2}. We compare our method against the standard CodeTree~\citep{li2024codetree} and Reflexion~\cite{shinn2023reflexion}, which prompts the LLM to repeatedly reflect on its previously generated code based on test case results. The evaluations are conducted with two LLMs including LLaMA-3.1-8B-Instruct, Qwen-2.5-Coder-1.5B-Instruct~\citep{hui2024qwen2}.

\vspace{-0.06in}
\subsection{Main Results.}\label{sec:main}
\vspace{-0.06in}
In this subsection, we present the main experimental results to demonstrate the effectiveness of DREAM in achieving a better accuracy-efficiency trade-off in reasoning.
The results for math reasoning tasks and code generation tasks are present in Section~\ref{sec:math_results} and~\ref{sec:code_results}, respectively.

\vspace{-0.06in}
\subsubsection{Math Reasoning}\label{sec:math_results}
\begin{figure}
    \centering
    \vspace{-0.2in}
    \includegraphics[width=1\linewidth]{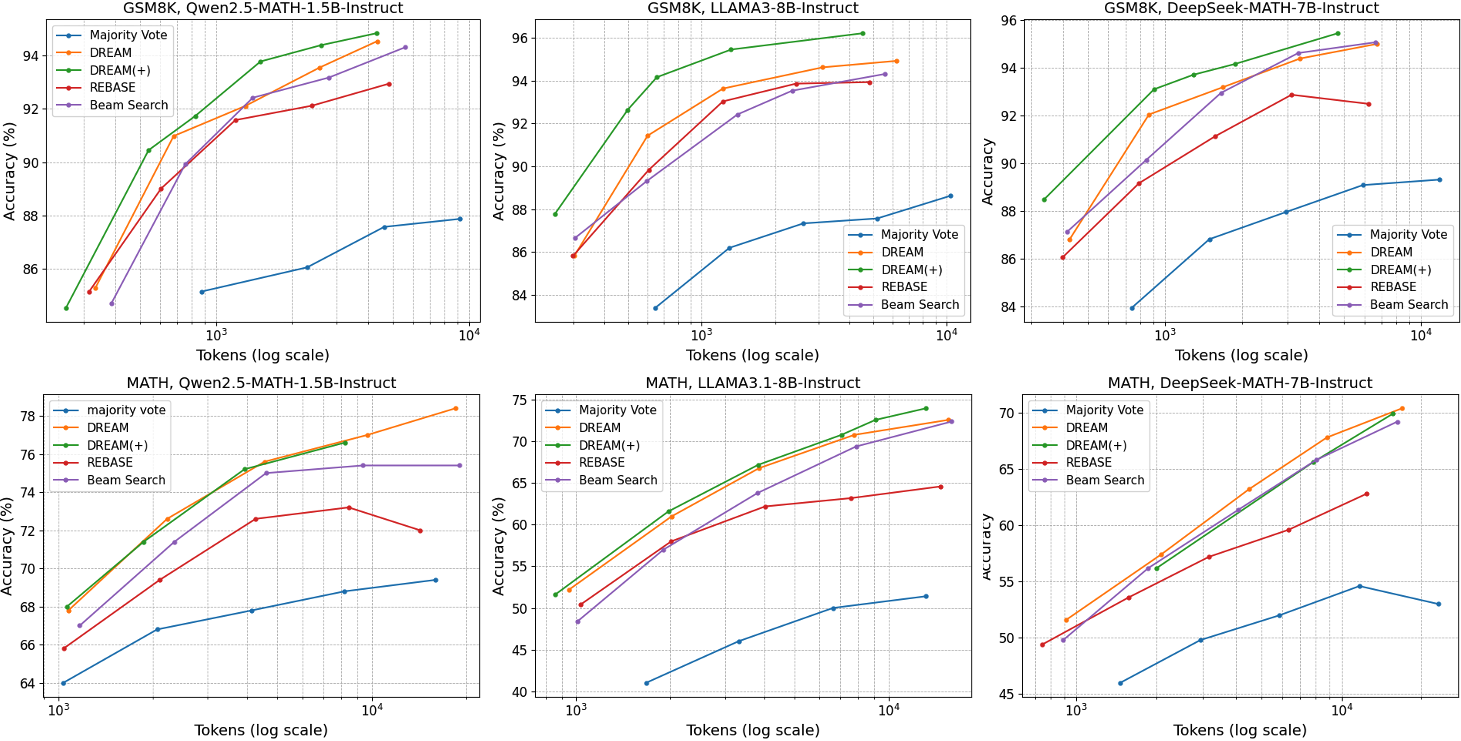}
     \vspace{-0.2in}
    \caption{\small{Accuracy vs Tokens (log) on GSM8K/MATH datasets}}\label{fig:main}
    \vspace{-0.1in}
\end{figure}

\vspace{-0.06in}
In this subsection, we present the main results of DREAM in math reasoning tasks. We show the accuracy–tokens frontier of each method in Figure~\ref{fig:main}. For DREAM, we consider the variants with/without budget allocation, which are labeled ``DREAM" and ``DREAM(+)", respectively.

Based on the results shown in Figure~\ref{fig:main}, we have the following observations. \textbf{First}, all tree-based search methods consistently achieve a significantly better accuracy–efficiency trade-off than majority vote. For example, on the MATH dataset with LLaMA-3.1-8B-Instruct, the performance gap can reach up to 20\%. This confirms the effectiveness of tree-structured search mechanisms in improving the accuracy–efficiency trade-off. In addition, the consistently strong performance across models also highlights the effectiveness of the reward model trained with the rollout-based annotation strategy.
\textbf{Second}, {in general, DREAM outperforms standard beam search which does not explicitly separate planning and execution in searching. For example, on the MATH dataset with Qwen2.5-MATH-1.5B, DREAM continues to outperform beam search, and the advantage becomes more pronounced as the token budget increases.
This suggests sampling and selecting planning and execution independently provides a more effective search of reasoning steps and leads to higher-quality candidate trajectories.}
\textbf{Third}, in many of our experiments, DREAM(+) with dynamic budget allocation provides additional gains in accuracy–efficiency trade-off, demonstrating the effectiveness of adaptively allocating computation based on the reward value. For example, on GSM8K with LLaMA-3-8B-Instruct, DREAM(+) consistently achieves about a 2\% improvement in accuracy over DREAM at comparable token budgets.
While in some cases (often on the MATH dataset), where the problems are more challenging, the improvement over standard dual-phase search is marginal, this can be explained by the fact that the adaptive budget mechanism is more effective when step difficulty is highly variable. When every step in a trajectory is uniformly hard, reallocating budget provides little benefit, and performance is mainly constrained by the inherent capacity of the reasoning model and the reward model. {This explanation is supported by the observation that in GSM8K, a large percentage (around 80\%) of reasoning steps trigger early stopping, whereas in MATH this percentage is much smaller (around 5\%). This suggests that GSM8K contains more diverse step-level difficulty, while MATH problems are more uniformly hard.}

\textbf{Finally}, we note that in many of our settings, the models used to develop the training data for the reward model are different from the models used in the final evaluation. {The only in-distribution setting is LLaMA-3-8B-Instruct with GSM8K, while all other evaluation settings are out of distribution. Nevertheless, in the out-of-distribution settings, DREAM consistently demonstrates strong performance,} indicating that our reward model generalizes well across different backbone LLMs rather than overfitting to the one used in training.

\vspace{-0.06in}
\subsubsection{Code Generation}
\vspace{-0.06in}
\label{sec:code_results}

\begin{figure}
    \centering
    \includegraphics[width=0.64\linewidth]{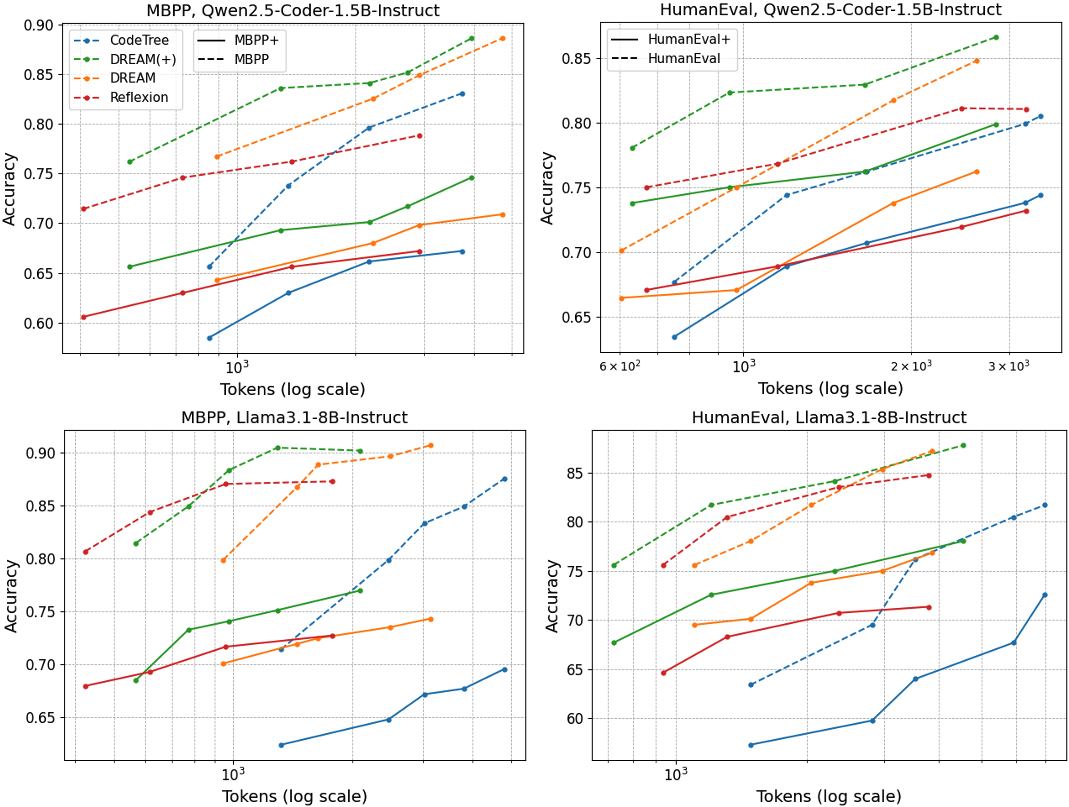}
    \vspace{-0.1in}
    \caption{\small{Accuracy vs Tokens (log) on MBPP/HumanEval datasets}}\label{fig:code}
     \vspace{-0.2in}
    \label{fig:placeholder}
\end{figure}

In this subsection, we present the main results of DREAM combined with CodeTree on code generation benchmarks, comparing with Reflexion and the standard CodeTree method. Figure~\ref{fig:code} shows the accuracy–token curve of each approach. Consistent with the math reasoning experiments in Section~\ref{sec:math_results}, we report two variants of DREAM: the version with/without budget allocation (denoted as ``DREAM/DREAM(+)'').

There are several observations from the experiment. \textbf{First, }according to the results in Figure~\ref{fig:code}, we observe that across all settings and datasets, the performance of CodeTree is significantly improved when combined with dual-phase search using reward models. For example, when the computation budget is around $10^3$ tokens, CodeTree+DREAM achieves an accuracy about 10\% higher than CodeTree on both MBPP and HumanEval with Qwen2.5-Coder-1.5B-Instruct. This demonstrates the effectiveness of separated searching of planning and execution and leveraging dedicated reward signals for each phase. In addition, applying budget allocation consistently provides further gains in the accuracy–efficiency trade-off. This suggests that adaptively allocating computation not only improves accuracy but also reduces unnecessary generation cost. \textbf{Second, }in some cases, when the token budget is small, Reflexion also achieves a strong accuracy–efficiency trade-off. For instance, with LLaMA-3.1-8B-Instruct, Reflexion performs better when the token budget (log scale) is below $10^3$. However, as the budget increases, the accuracy of Reflexion grows more slowly compared to CodeTree+DREAM, indicating that Reflexion saturates earlier, while dual-phase search continues to benefit from additional computation. \textbf{Finally}, it is also worth noting that HumanEval does not appear in the training data of the reward model. 
{The advantage of CodeTree+DREAM in Figure~\ref{fig:code} also demonstrates the strong generalization ability of our reward model to unseen datasets.}

In short, these findings highlight the advantages of integrating dual-phase search and adaptive budget allocation into tree-based code generation framework.

\vspace{-0.06in}
\subsection{Generalization of the Math Reward Model to Unseen Datasets}
\vspace{-0.06in}
In the main results, we demonstrated the transferability of the reward model in the code generation task on unseen datasets, as the HumanEval dataset does not appear in the training data of the reward model. In this section, we further examine the reward model's transferability in the math reasoning domain. Specifically, we consider two out-of-distribution datasets: AMC23~\citep{mathai_amc23}, which consists of 40 competition-style problems from the American Mathematics Competitions 2023, and the test set of ASDiv~\citep{miao2021diverse}, which contains 301 diverse grade-school math word problems. We evaluate reasoning with DREAM/DREAM(+) using LLaMA3/3.1-8B-Instruct, and compare it with a simple majority vote baseline. The results are presented in Table~\ref{tab:ood-math}.

\begin{table}[h!]
 \vspace{-0.2in}
\centering 
\caption{\small{Performance in out-of-distribution datasets} }\label{tab:ood-math}
\resizebox{0.9\textwidth}{!}{
\begin{tabular}{ll ll ll |ll ll ll}
\toprule
\multicolumn{6}{c|}{AMC23 (LLaMA3.1)} & \multicolumn{6}{c}{ADSIV (LLaMA3)} \\
\midrule
\multicolumn{2}{c}{DREAM(+)} & \multicolumn{2}{c}{DREAM} &
\multicolumn{2}{c}{Majority Vote} &
\multicolumn{2}{c}{DREAM(+)} & \multicolumn{2}{c}{DREAM} &
\multicolumn{2}{c}{Majority Vote} 
\\
\cmidrule(r){1-2} \cmidrule(r){3-4} \cmidrule(r){5-6}\cmidrule(r){7-8} \cmidrule(r){9-10}  \cmidrule(r){11-12} 
 acc & \# tokens & acc & \# tokens & acc & \# tokens &  acc &  \# tokens&  acc &  \# tokens &  acc &  \# tokens \\
\midrule
37.50\%	&3186.38& 30.00\% & 3149.6  & 22.50\% & 2851.45& 95.35\%	& 160.55 & 93.02\%	& 162.65  & 93.02\% & 193.93\\
47.50\%	& 6068.18& 42.50\% & 6256.20 & 22.59\% & 5469.45 &96.35\%&	218.76 & 95.35\%&332.60 & 94.02\% & 388.29\\
 50.00\%&	11210.10 & 47.50\% & 12430.3  &  27.50\% &11708.9& 97.67\%&	543.02 & 96.01\%&650.10 & 95.35\% & 782.80\\
60.00\%	& 23515.20 & 50.00\%&  22971.6 & 30.00\% &22586.3& 98.01\%&	1112.1 &97.34\% & 1314.7& 95.02\% &1558.1\\
\bottomrule
\end{tabular}} 
 \vspace{-0.1in}
\end{table}

From Table~\ref{tab:ood-math}, we observe that across both datasets, DREAM with our reward model consistently achieves higher accuracy at the same level of tokens compared to majority vote, and DREAM(+) provide additional benefits in terms of accuracy-efficiency trade-off.
Notably, on AMC23, which is significantly more challenging than the datasets used to train the reward model, our approach still provides strong guidance for intermediate reasoning, showing up to a 30\% improvement over majority voting.
These findings suggest that our reward model generalizes beyond its training distribution and demonstrates strong transferability to out-of-distribution math reasoning tasks.
\vspace{-0.06in}
\subsection{Ablation Studies}\label{sec:abla}

In this section, we conduct ablation studies regarding the reward model size for math reasoning (Section~\ref{sec:rewardsize}), the effect of using the critic agent in code generation task and discuss the synergy effect of DREAM. Due to space limitations, we defer the latter two studies to Appendix~\ref{appd:critic} and~\ref{appd:synergy}.
\subsubsection{Reward Model Size}\label{sec:rewardsize}
\vspace{-0.06in}
In this subsection, we study the impact of reward model size. In our main experiments for math reasoning, the reward model is fine-tuned from Qwen2.5-32B-Instruct, which is relatively large. To assess whether smaller models can serve as effective alternatives, we additionally fine-tune a reward model based on Qwen2.5-7B-Instruct. We then compare the performance of standard dual-phase search (without budget allocation) under two configurations: using Qwen2.5-7B-Instruct or Qwen2.5-32B-Instruct as the reward model. For reference, we also report results with simple majority voting. We conduct the experiments using LLaMA3/3.1-8B-Instruct as the reasoning models and the results are summarized in Table~\ref{tab:size}.

\begin{table}[h!]
\vspace{-0.13in}
\centering 
\caption{Performance comparison across different reward model size. }\label{tab:size}
\resizebox{1.0\textwidth}{!}{
\begin{tabular}{ll ll ll |ll ll ll}
\toprule
\multicolumn{6}{c|}{GSM8K} & \multicolumn{6}{c}{MATH} \\
\midrule
\multicolumn{2}{c}{Qwen2.5-32B} &
\multicolumn{2}{c}{Qwen2.5-7B} &
\multicolumn{2}{c|}{Majority vote} & \multicolumn{2}{c}{Qwen2.5-32B} &
\multicolumn{2}{c}{Qwen2.5-7B} &
\multicolumn{2}{c}{Majority vote}\\
\cmidrule(r){1-2} \cmidrule(r){3-4} \cmidrule(r){5-6}\cmidrule(r){7-8} \cmidrule(r){9-10} \cmidrule(r){11-12}
 acc & \# tokens & acc & \# tokens & acc & \# tokens &  acc & \# tokens & acc & \# tokens & acc & \# tokens\\
\midrule
 91.43\% & 601.13 & 87.34\% & 606.191  &83.40\% &646.18 & 61.00\% & 2021.16& 55.20\% & 1829.73 &41.00\% & 1674.98 \\
   93.63\% &  1221.21 & 93.63\% & 1191.49&86.20\%&1297.45 &	66.80\% &3854.31 &58.40\%&	3576.18	 & 46.00\% & 3314.41 \\
 94.62\% & 2463.82 & 94.62\% & 2385.57 &  87.34\% &2598.09 & 70.80\%	&7776.58&	62.80\%	&7477.92 &50.00\%& 6647.12 \\
94.92\% & 4916.89 & 94.92\% & 4444.57  & 87.57\% & 5202.77 & 72.60\%	&15553.2	&65.60\%&	14557.8 & 51.40\% &  13157.1\\
\bottomrule
\vspace{-0.27in}
\end{tabular}} 
\end{table}

According to the results, the 32B reward model consistently outperforms the 7B version, achieving better accuracy-efficiency trade-offs across datasets. This suggests that, as larger instruction-tuned LLMs possess stronger intrinsic reasoning and representation capabilities, fine-tuning them with rollout-based supervision enables the reward function to better capture nuanced signals of correctness. Nevertheless, the 7B reward model also demonstrates strong effectiveness: although its step selection is not as precise as the 32B model, it still achieves significantly better accuracy–efficiency trade-offs compared to majority voting. This suggests that, with properly labeled training data, even moderately sized reward models can provide substantial benefits while reducing computations.


\vspace{-0.1in}
\section{Conclusion}
\vspace{-0.06in}

In this work, we proposed DREAM, a dual-phase test-time scaling framework that explicitly separates reasoning into planning and execution and conducts dedicated search in each phase. By equipping each phase with a reward model and introducing an adaptive budget allocation mechanism, our method enables finer-grained control over reasoning search, reduces wasted computation, and improves accuracy on both math reasoning and code generation tasks. Empirical results show that DREAM consistently outperforms standard beam search and prior PRM-based methods, highlighting the benefit of searching planning and execution separately and adaptively managing budget.






\bibliography{conference}
\bibliographystyle{conference}
\newpage
\appendix
\section{Appendix}
\subsection{Algorithm of DREAM(+)}\label{appd:alg}
We present the detailed algorithm of DREAM(+) in Algorithm~\ref{alg}.
\begin{algorithm}
\caption{Dual-Phase Search with budgets adjustment.}
\label{alg}
\begin{algorithmic}[1]
\Require Question $Q$, Planning/Execution Budget $N_1$/$N_2$, Thresholds $\tau_{p_1}$/ $\tau_{p_2}$ and $\tau_{e_1}$ /$\tau_{e_2}$, Beam Width $n_1$/$n_2$, Additional Budget limit $m_1$/$m_2$.
\State Initialize finished paths $F \gets \emptyset$
\State Initialize step counter $s \gets 1$
\State Initialize beam set $B \gets \emptyset$
\While{$s \leq$ max\_steps}
    \If{ all $b \in B$ is finished} \State \textbf{break} \EndIf

    \State $C_{p} (\text{candidates storage}) \gets \emptyset$
        \If{$s = 1$}     
        \State Generate up to $N_1$ candidates for planning $p_1$ with reward scores $r$
        \State Early stop if there are $n_1$ planning all having reward $r > \tau_{p_1}$
         \State $C_p \gets C_p \cup \{(p_1^{(i)},r_1^{(i)}) \mid i = 1, \dots, n\}$, $n$ denotes the number of actual sampling
        \If{all $r < \tau_{p2}$ }
            \State Generate up to additional $m_1$ candidates for planning $p_1$  with reward scores $r$
            \State Early stop if there are $n_1$ planning having $r > \tau_{p1}$
            \State $C_p \gets C_p \cup \{(p_1^{(i)},r_1^{(i)}) \mid i = 1, \dots, n\}$ 
        \EndIf
        \Else
            \ForAll{$b \in B$}
                \State Generate up to $N_1 / n_1$ candidates for planning $p_s$ with reward scores $r$
                \State Early stop if there are $N_1 / n_1$ planning having $r > \tau_{p1}$
                 \State $C_p \gets C_p \cup \{(p_s^{(i)},r_s^{(i)}) \mid i = 1, \dots, n\}$
                 \If{all $r < \tau_{p2}$ }
                \State Generate up to $m_1$ additional candidates for planning $p_s$ with reward scores $r$
                \State Early stop if there are $N_1 / n_1$ planning having $r > \tau_{p1}$
                 \State $C_p \gets C_p \cup \{(p_s^{(i)},r_s^{(i)}) \mid i = 1, \dots, n\}$
                  \EndIf
            \EndFor
        \EndIf
        \State Sort $C_p$ and take the top-$n_1$ planning $\{(p_1^1,e_1^1, \cdots, p_{s-1}^1, e_{s-1}^1,p_s^1, r_s^1), \cdots, (p_1^{n_1},e_1^{n_1}, \cdots, p_{s-1}^{n_1}, e_{s-1}^{n_1},p_s^{n_1}, r_s^{n_1}) \}$
        \State Update beam $B \gets \{(Q, p_1^1, e_1^1, \cdots ,p_{s-1}^1, e_{s-1}^1, p_s^1 ), \cdots, (Q, p_1^{n_1}, e_1^{n_1}, \cdots,  p_{s-1}^{n_1}, e_{s-1}^1, p_s^{n_1} )\}$
         \State $C_{e} (\text{candidates storage}) \gets \emptyset$
        \ForAll{$b \in B$}
        \State Generate up to $N_2 / {n_2}$ candidates for  executions $e_s$ based on $p_s$ with reward scores $r'$
        \State Early stop if there are $N_2 / {n_2}$  executions having reward $r' > \tau_{e1}$
         \State $C_e \gets C_e \cup \{(p_s^{(i)}, e_s^{(i)},r{'}_s^{(i)}) \mid i = 1, \dots, n\}$, $n$ denotes the number of actual sampling
            \If{all $r' < \tau_2$  }
        \State Generate up to $m_2$ additional candidates for executions $e_s$ with reward scores $r'$
        \State Early stop if there are $N_2 / {n_2}$  executions having reward $r' > \tau_{e1}$
         \State $C_e \gets C_e \cup \{(p_s^{(i)}, e_s^{(i)},r{'}_s^{(i)}) \mid i = 1, \dots, n\}$
            \EndIf
        \EndFor
           \State Sort $C_e$  and take the top-$n_2$  executions $\{(p_1^1,e_1^1, \cdots,p_s^1,e_s^1, {r'}_s^1), \cdots, (p_1^{n_2},e_1^{n_2}, \cdots,p_s^{n_2}, e_s^{n_2}, {r'}_s^{n_2}) \}$
        \State Update beam $B \gets \{(Q, p_1^1, e_1^1, \cdots ,p_s^1,e_s^1 ), \cdots, (Q, p_1^{n_2}, e_1^{n_2}, \cdots,p_s^{n_2}, e_s^{n_2} )\}$
    \State $s \gets s + 1$
\EndWhile
\State \Return Path with highest reward in $B$
\end{algorithmic}
\end{algorithm}

\subsection{Examples to demonstrate difficulty variants in reasoning steps.}\label{appd:examp}
In this subsection, we present real examples to illustrate difficulty diversity both across problems in a dataset and across steps within a single problem, motivating the need for dynamic allocation. Tables~\ref{tab:ex_gsm8k} and \ref{tab:ex_math} provide examples from GSM8K and MATH: for each dataset, we include one easy problem, where intermediate steps have high average reward values and low variance, and one hard problem, where intermediate steps have lower average reward values and higher variance. For reference, we also provide the ground-truth solution for each example, offering a more intuitive sense of the difficulty of each problem. These examples demonstrate that difficulty variance arises not only across problems but also within individual reasoning trajectories.

\begin{table}[h!]
\centering 
\caption{Examples of samples with different diversity in GSM8K datasets}\label{tab:ex_gsm8k}
\resizebox{1.0\textwidth}{!}{
\begin{tabular}{l}
\toprule
\midrule
Q: Josh decides to try flipping a house.  He buys a house for \$80,000 and then puts in \$50,000 in repairs.  \\This increased the value of the house by 150\%.  How much profit did he make? \\\\
Ground truth solution:\\
The cost of the house and repairs came out to $80,000+50,000=130,000$\\
He increased the value of the house by $80,000*1.5=120,000$\\
So the new value of the house is $120,000+80,000=200,000$ \\
So he made a profit of $200,000-130,000=70000$\\
\#\#\#\# 70000 \\ \\
Average reward score for intermediate steps: 0.3613\\
Standard deviation of reward score: 0.184845\\
\midrule
Q: A robe takes 2 bolts of blue fiber and half that much white fiber.  How many bolts in total does it take? \\ \\
Ground truth solution: \\
It takes $2/2=1$ bolt of white fiber \\
So the total amount of fabric is $2+1=3$ bolts of fabric \\
\#\#\#\# 3 \\
\\
Average reward score for intermediate steps: 0.9990\\
Standard deviation of reward score: 0.001746 \\
\bottomrule
\vspace{-0.3in}
\end{tabular}} 
\end{table}

\begin{table}[h!]
\centering 
\caption{Examples of samples with different diversity in MATH datasets}\label{tab:ex_math}
\resizebox{1.0\textwidth}{!}{
\begin{tabular}{l}
\toprule
\midrule
Q: What is the smallest positive integer $n$ such that all the roots of $z^4 + z^2 + 1 = 0$ are $n^{\text{th}}$ roots of unity?\\\\
Ground Truth:\\ 
Multiplying the equation $z^4 + z^2 + 1 = 0$ by $z^2 - 1 = (z - 1)(z + 1)$, we get $z^6 - 1 = 0$. \\
Therefore, every root of $z^4 + z^2 + 1 = 0$ is a sixth root of unity.\\
The sixth roots of unity are $e^{0}$, $e^{2 \pi i/6}$, $e^{4 \pi i/6}$, $e^{6 \pi i/6}$, $e^{8 \pi i/6}$, and $e^{10 \pi i/6}$. \\
We see that $e^{0} = 1$ and $e^{6 \pi i/6} = e^{\pi i} = -1$, 
so the roots of $z^4 + z^2 + 1 = 0$ \\ are the remaining sixth roots of unity, namely $e^{2 \pi i/6}$, $e^{4 \pi i/6}$, $e^{8 \pi i/6}$, and $e^{10 \pi i/6}$. \\
The complex number $e^{2 \pi i/6}$ is a primitive sixth root of unity, so by definition, \\ the smallest positive integer $n$ such that $(e^{2 \pi i/6})^n = 1$ is 6. \\
Therefore, the smallest possible value of $n$ is $\boxed{6}$.
\\ \\
Average reward score for intermediate steps: 0.4206503 \\
Standard deviation of the reward score: 0.3246094 \\
\midrule
Q: Simplify $\sqrt{242}$.    \\\\  
Ground truth:      \\ 
Factor 242 as $11^2 \cdot 2$.  \\
Then $\sqrt{242} = \sqrt{11^2} \cdot \sqrt2 = \boxed{11\sqrt2}$. \\
\\
Average reward score for intermediate steps: 0.9947\\
Standard deviation of reward score: 0.0024\\
\bottomrule
\vspace{-0.3in}
\end{tabular}} 
\end{table}

\section{Additional Details of Experiments Setting}\label{appd:setting}

In this subsection, we provide additional details of experiment settings to ensure reproducibility.
\begin{itemize}
\item For math reasoning, we develop the reward model by fine-tuning \text{Qwen2.5-32B-Instruct} for 2 epochs with a learning rate of $2.0 \times 10^{-6}$, using the Adam optimizer and a cosine learning-rate scheduler. The fine-tuning is conducted on 8 H100 GPUs with a batch size of 32. For code generation, we fine-tune \text{Qwen2.5-Coder-7B-Instruct} for 3 epochs, while keeping the other hyperparameters the same.
\item During inference on math reasoning, we set the sampling temperature to 1.0 for \text{LLaMA3/3.1-8B-Instruct} and \text{DeepSeek-MATH-7B-Instruct}, and 0.5 for \text{Qwen2.5-MATH-1.5B-Instruct}.
\item For code generation, we set the temperature to 1.0 across all experiments in DREAM and DREAM(+), while using 0.0 for \text{CodeTree} and \text{Reflexion}.
\end{itemize}

\vspace{-0.06in}
\section{Additional Ablation Studies}

\subsection{Usage of Critic Agents in Code Generation}\label{appd:critic}
\vspace{-0.06in}
In this subsection, we provide empirical evidence showing that when applying dual-phase search (DREAM) in the CodeTree method, the critic agent does not yield clear benefits in performance and, in fact, reduces efficiency. Table~\ref{tab:critic} presents results for DREAM+CodeTree under settings with and without critic agents. The accuracies on MBPP/HumanEval and MBPP+/HumanEval+ are reported as “weak acc” and “acc,” respectively.

\begin{table}[h!]
\centering 
\vspace{-0.1in}
\caption{Comparison of performance with/without critic agent }\label{tab:critic}
\resizebox{1.01\textwidth}{!}{
\begin{tabular}{lll |lll |lll |lll }
\toprule
\multicolumn{6}{c|}{MBPP(+)} & \multicolumn{6}{c}{HumanEval(+)} \\
\midrule
\multicolumn{3}{c|}{without critic} &
\multicolumn{3}{c|}{with critic} &
\multicolumn{3}{c|}{without critic} &
\multicolumn{3}{c}{with critic} 
\\
\cmidrule(r){1-3} \cmidrule(r){4-6} \cmidrule(r){7-9}\cmidrule(r){10-12} 
 weak acc & acc  & \# tokens &  weak acc & acc  & \# tokens &  weak acc & acc  & \# tokens &  weak acc & acc  & \# tokens \\
\midrule
81.49\%	& 68.52\%	& 569.59  & 81.49\%	& 68.52\%	& 1051.52& 75.61\%	& 67.69\%	& 721.28& 75.61\%	&67.69\%	&1235.92 \\
86.51\%	& 74.10\%	& 771.66 & 87.58\%&	74.88\% &	1348.16   & 81.71\%& 	72.57\%& 	1205.10 & 78.66\%	& 70.13\%	& 1596.59\\
88.37\% & 	74.09\% &	976.72& 86.25\% &	73.02\%	& 1627.35 &	84.14\%  & 72.57\% & 	1619.51 & 80.49\%&	71.34\% &	1827.49 \\
90.48\%&	75.14\%	& 1292.83 & 88.37\%	& 76.19\%	& 2074.12 & 84.14\% & 	75.00\%& 	2307.87 &81.71\% &	71.95\%	 & 2311.83 \\
90.21\% &	76.98\% &	2081.73 & 90.73\%&	75.94\%	&2620.38 & 87.80\%&	78.04\% & 	4552.75 & 87.20\%&	76.83\%& 	4833.02 \\
\bottomrule
\end{tabular}} 
\end{table}

From the results, we observe that to achieve the same level of reasoning accuracy, DREAM with a critic agent consistently requires substantially more generated tokens. This indicates that, although the critic agent proposed by \citet{li2024codetree} was originally shown to improve  accuracy given sufficient computation budget, its efficiency is inferior to simply scaling generation multiple times and directly applying the percentage of passed test cases as the reward signal. These findings support our design choice of removing the critic agent from the code generation framework, simplifying the system while preserving performance and improving efficiency.

\subsection{Synergy Effect of DREAM}\label{appd:synergy}
\begin{figure}[H]
    \centering
    \includegraphics[width=0.5\linewidth]{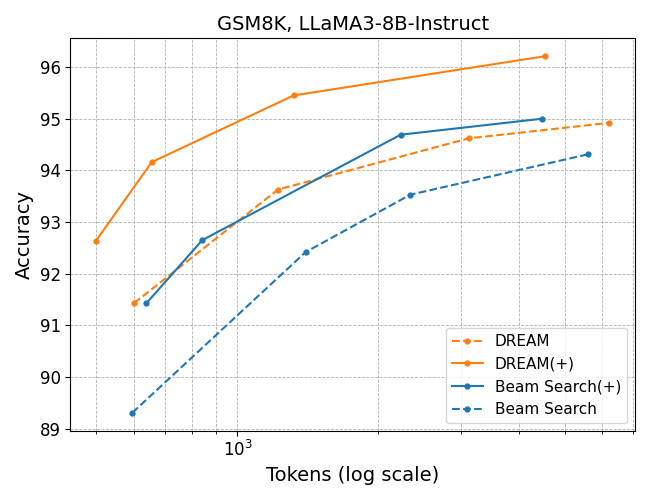}
    \caption{Accuracy-Tokens Frontier across methods.}
    \label{fig:synergy}
\end{figure}

In this subsection, we demonstrate the synergy between dual-phase search and dynamic budget allocation. Specifically, we compare the benefits of applying budget allocation to dual-phase search versus standard beam search. Figure~\ref{fig:synergy} plots the accuracy–tokens frontier of four methods: DREAM, DREAM(+), Beam Search, and Beam Search(+), where the “+” variants denote the incorporation of dynamic budget allocation. From the figure, we observe that budget allocation also improves the accuracy–efficiency trade-off when applied to standard beam search, indicating the general effectiveness of this design. However, the gains for Beam Search(+) are consistently smaller than those achieved by DREAM(+). This comparison highlights the complementary nature of the two components: dual-phase search and dynamic budget allocation. The synergy arises because dual-phase search reduces wasted computation by separating planning and execution, while dynamic budget allocation further reallocates saved resources to harder steps, making the two mechanisms mutually reinforcing.


\end{document}

%% file: math_commands.tex
\usepackage{amsmath,amsfonts,bm}

\def\eqref#1{equation~\ref{#1}}

\def\1{\bm{1}}

\DeclareMathAlphabet{\mathsfit}{\encodingdefault}{\sfdefault}{m}{sl}
\SetMathAlphabet{\mathsfit}{bold}{\encodingdefault}{\sfdefault}{bx}{n}

